\def\mode{icpm}
\def\icpm{%
\documentclass[conference]{./IEEEtran}		
\bibliographystyle{./IEEEtran}
}
\def\article{%
\documentclass{article}	 
\setlength{\paperheight} {232.8mm}
\setlength{\paperwidth}  {151.5mm}
\setlength\voffset       {-23mm}
\setlength\hoffset       {-34mm}
\bibliographystyle{plain}
}
\csname\mode\endcsname

\newcommand{\articletitle}   {An Entropic \Relevance\ Measure for Stochastic Conformance Checking in Process Mining}
\newcommand{\articlesubjet}  {Computer Science, Process Mining, Process Querying, Information Systems}

\newcommand{\articleauthors} {Artem~Polyvyanyy, Alistair~Moffat, Luciano~Garc{\'{\i}}a{-}Ba{\~{n}}uelos}

\newcommand{\Relevance}	{{Relevance}} 
\newcommand{\relevance}	{{relevance}}

\newif\ifshowtodos
\showtodosfalse

\usepackage[usenames,dvipsnames]{xcolor}
\usepackage[algoruled,vlined,linesnumbered]{algorithm2e}
\usepackage{amsmath}
\usepackage{amsfonts}
\usepackage{mdframed}
\usepackage{paralist}
\setdefaultleftmargin{1em}{1em}{1em}{1em}{1em}{1em}
\usepackage{datetime}
\newdateformat{monthyeardate}{\monthname[\THEMONTH] \THEYEAR}
\usepackage[pdfsubject={\articlesubjet},
                            pdftitle={\articletitle},
                            pdfauthor={\articleauthors},
                            pdfkeywords={\articlesubjet},
                            colorlinks=false,
                            pdfborder={0.25 0.25 0.25}]{hyperref}
\hypersetup{bookmarksdepth}
\usepackage[algoruled]{algorithm2e}
\usepackage{tikz,pgf,xcolor}
\usetikzlibrary{arrows,arrows.meta,petri,backgrounds,snakes,automata,trees,plotmarks,shadows,shapes,positioning}
\usepackage{pgfplots}
\pgfplotsset{compat=1.5}
\usetikzlibrary{pgfplots.groupplots}
\usepackage{nicefrac}
\usepackage{url}
\usepackage{mathtools}
\usepackage[firstpage]{draftwatermark}
\SetWatermarkAngle{0}
\SetWatermarkFontSize{12pt}
\SetWatermarkHorCenter{178.5mm}
\SetWatermarkVerCenter{264mm}
\SetWatermarkLightness{0.6}
\SetWatermarkText{Postprint{,} \monthyeardate\today}
\usepackage[capitalise,nameinlink]{cleveref}
\usepackage{subfig}
\definecolor{mybluecolor}{RGB}{50,106,218}
\definecolor{myredcolor}{RGB}{176,53,53}
\definecolor{mygreencolor}{RGB}{93,172,0}
\definecolor{myyellowcolor}{RGB}{255,163,34}
\definecolor{mypurplecolor}{RGB}{86,35,132}
\definecolor{mytealcolor}{RGB}{30,161,165}

\newcommand{\splitatcommas}[1]{%
  \begingroup
  \ifnum\mathcode`,="8000
  \else
    \begingroup\lccode`~=`, \lowercase{\endgroup
      \edef~{\mathchar\the\mathcode`, \penalty0 \noexpand\hspace{-1pt plus 1em}}%
    }\mathcode`,="8000
  \fi
  #1%
  \endgroup
}

\newcommand{\mult}{{\ensuremath \,}}

\newcommand{\func}[3]{{{#1}:{#2} \rightarrow {#3}}}
\newcommand{\funcCall}[2]{{\ensuremath {\mathit{#1}}_{\!}\left({#2}\right)}}

\newcommand{\intintervalcc}[2]{{\ensuremath \left[#1 \,..\, #2\right]}}
\newcommand{\intervalcc}[2]{{\ensuremath \left[#1, #2\right]}}

\providecommand{\cardinality}[1]{\ensuremath \left|{#1}\right|}
\renewcommand{\cardinality}[1]{\ensuremath \left|{#1}\right|}
\providecommand{\multiplicity}[2]{\ensuremath \funcCall{m_{#1}}{#2}}
\renewcommand{\multiplicity}[2]{\ensuremath \funcCall{m_{#1}}{#2}}
 \newcommand{\mset}[1] {\ensuremath [\splitatcommas{#1}]}
\newcommand{\msetel}[2]{{\ensuremath {{#1}^{#2}}}}

\newcommand{\kleenestar}[1]{{\ensuremath {#1}^{*}}}
\newcommand{\emptysequence}{{\ensuremath \epsilon}}
\newcommand{\sequence}[1]{\ensuremath \langle\splitatcommas{#1}\rangle}
\newcommand{\seqLength}[1]{\ensuremath \left|{#1}\right|}
\newcommand{\concat}[2]{\ensuremath #1 \circ #2}

\newcommand{\set}[1]{\ensuremath \{\splitatcommas{#1}\}}
\newcommand{\setbuilder}[2]{\ensuremath \{ #1 \,|\, #2 \}}
\newcommand{\setCardinality}[1]{\ensuremath \left|{#1}\right|}

\newcommand{\pair}[2]{\ensuremath (\splitatcommas{#1, #2})}
\newcommand{\tuple}[1]{\ensuremath \left(\splitatcommas{#1}\right)}

\newcommand{\fig}[9]{\begin{figure}[#1]
\vspace{#2mm}
\begin{center}
	\includegraphics[scale=#3,trim=#4]{#5}
\end{center}
\vspace{#6mm}
\caption{\small #7.}
\vspace{#8mm}
\label{#9}
\end{figure}}

\newcommand{\relplot}[1]{\begin{tikzpicture}
	\begin{axis}[
	  legend columns = 4,
		legend to name = rellegend,
	  xlabel style={font=\scriptsize},
		ylabel style={font=\scriptsize},
	  height=4.2cm,
		width=.37\linewidth,
		label style={font=\scriptsize},
		xlabel={\scriptsize Size},
		ylabel={\scriptsize Bits},
		xlabel shift = -.15cm,
		ylabel shift = -.1cm,
		mark options={solid},
		ylabel near ticks,
		xlabel near ticks,
		x tick label style={/pgf/number format/1000 sep=, font=\tiny},
		y tick label style={/pgf/number format/1000 sep=, font=\tiny},
		legend style={at={(0.48,-0.25)}, anchor=north, font=\scriptsize},
		xmin=0,
		xmax=,
		ymin=0,
		grid=both
		]
		
    \pgfplotstableread[col sep=comma]{#1}{\table}
				
		\addplot[densely dashdotted, ultra thick, mark=pentagon*, mark size=0.75pt, mark options={solid}, color=alistairyellowcolor] table [y index=4,col sep=comma] {\table}; 
		\addlegendentryexpanded{Selector coding cost}
		
		\addplot[densely dashed, ultra thick, mark=square*, mark size=0.75pt, mark options={solid}, color=alistairbluecolor] table [y index=2,col sep=comma] {\table}; 
    \addlegendentryexpanded{Background coding cost}
		
		\addplot[densely dotted, ultra thick, mark=diamond*, mark size=0.75pt, mark options={solid}, color=alistairgreencolor] table [y index=3,col sep=comma] {\table}; 
    \addlegendentryexpanded{Model coding cost}
		
		\addplot[solid, ultra thick, mark=*, mark size=0.75pt, mark options={solid}, color=alistairredcolor] table [y index=1,col sep=comma] {\table};	
    \addlegendentryexpanded{Entropic {\relevance}}
	\end{axis}
\end{tikzpicture}}

\newcommand{\scatterplot}[3]{\begin{tikzpicture}
	\begin{axis}[
	  legend columns = 4,
		legend to name = relscatterlegend,
	  xlabel style={font=\scriptsize},
		ylabel style={font=\scriptsize},
	  height=4.2cm,
		width=.37\linewidth,
		label style={font=\scriptsize},
		xlabel={\scriptsize Size},
		ylabel={\scriptsize Bits},
		xlabel shift = -.15cm,
		ylabel shift = -.1cm,
		mark options={solid},
		ylabel near ticks,
		xlabel near ticks,
		x tick label style={/pgf/number format/1000 sep=, font=\tiny},
		y tick label style={/pgf/number format/1000 sep=, font=\tiny},
		legend style={at={(0.48,-0.25)}, anchor=north, font=\scriptsize},
		xmax=,
		grid=both
		]
		
    \pgfplotstableread[col sep=comma]{#1}{\tablecelonispe}
		\pgfplotstableread[col sep=comma]{#2}{\tablecelonisve}
		\pgfplotstableread[col sep=comma]{#3}{\tablesander}
		
		\addplot[mark=diamond*, only marks, mark size=1.5pt, color=myorangecolor] table [y index=1,col sep=comma] {\tablesander}; 
    \addlegendentryexpanded{DFM\,\,\,}
		
		\addplot[mark=*, only marks, mark size=1.5pt, color=mypurplecolor] table [y index=1,col sep=comma] {\tablecelonispe};	
    \addlegendentryexpanded{Celonis PE\,\,\,}
				
		\addplot[mark=square*, only marks, mark size=1.5pt, color=mytealcolor] table [y index=1,col sep=comma] {\tablecelonisve}; 
    \addlegendentryexpanded{Celonis VE}
		
	\end{axis}
\end{tikzpicture}}

\newcommand{\orcidartem}		{\href{https://orcid.org/0000-0002-7672-1643}{\protect\includegraphics[scale=0.05]{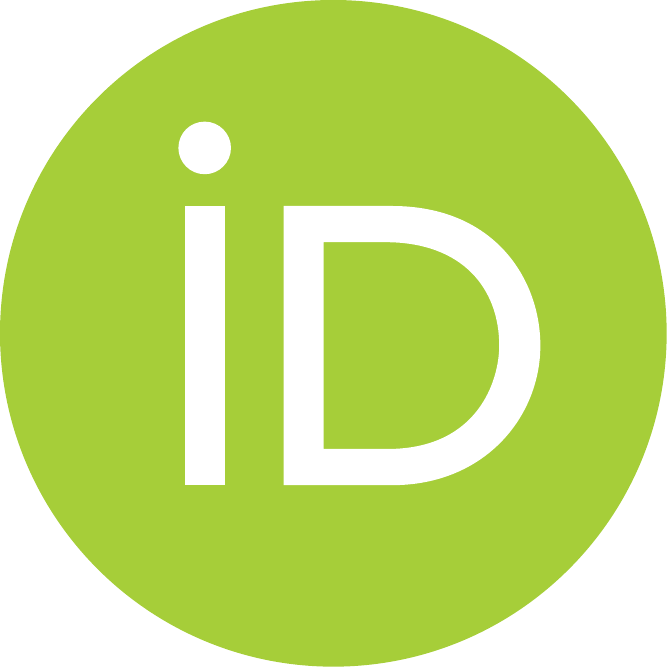}}}
\newcommand{\orciluciano}		{\href{https://orcid.org/0000-0001-9076-903X}{\protect\includegraphics[scale=0.05]{fig/orcid}}}

\newcommand{\actions}			{{\ensuremath \Lambda}}

\newcommand{\natnumwithzero}		{\ensuremath \mathbb{N}_0}
\newcommand{\nonnegrealnumbers}	{\ensuremath \mathbb{R}^+}

\newcommand{\ie}					{i.e.,}

\newcommand{\iffs}				{iff~}

\definecolor{mybluecolor}	 {RGB}{50,106,218}
\definecolor{myredcolor} 	 {RGB}{176,53,53}
\definecolor{mygreencolor} {RGB}{93,172,0} 
\definecolor{myorangecolor}{RGB}{255,163,34}
\definecolor{mybrowncolor} {RGB}{154,3,30} 
\definecolor{myorangecolor2}{RGB}{238,108,52} 
\definecolor{mybluecolor2}	{RGB}{0,78,152} 
\definecolor{mypurplecolor}{RGB}{86,35,132}
\definecolor{mytealcolor}{RGB}{30,161,165}

\definecolor{alistairbluecolor}	 {RGB}{0,0,209}
\definecolor{alistairredcolor} 	 {RGB}{212,19,19}
\definecolor{alistairgreencolor} {RGB}{0,175,0} 
\definecolor{alistairpurplecolor}{RGB}{142,0,142}
\definecolor{alistairyellowcolor}{RGB}{255,214,0}

\newtheorem{mytheorem}		{Theorem}
\newtheorem{mydefinition}	{Definition}
\newtheorem{mylemma}			{Lemma}
\newtheorem{myproposition}{Proposition}
\newtheorem{mycorollary}	{Corollary}
\newtheorem{myexample}		{Example}
\newtheorem{myconjecture}	{Conjecture}

\newtheorem{myinvariant}	{Invariant}

\numberwithin{mytheorem}		{section}
\numberwithin{mydefinition}	{section}
\numberwithin{mylemma}			{section}
\numberwithin{myproposition}{section}
\numberwithin{mycorollary}	{section}
\numberwithin{myexample}		{section}
\numberwithin{myconjecture}	{section}
\numberwithin{myremark}			{section}
\numberwithin{myinvariant}	{section}

\newenvironment{define}[3][]
{\begin{mydefinition}[#2]\label{#3}#1\normalfont}
{\hfill\ensuremath{\lrcorner}\end{mydefinition}}

\addtocounter{mytheorem}{0}

\newcommand{\orcidalistair}		{\href{https://orcid.org/0000-0002-6638-0232}{\protect\includegraphics[scale=0.05]{fig/orcid}}}

\newcommand{\etal}{{et al.}}
\newcommand{\bits}{\mbox{bits}}
\newcommand{\trace}{\mbox{$t$}}
\newcommand{\strace}{\mbox{\scriptsize$t$}}

\title{\articletitle}

\date{\today}

\author{
\IEEEauthorblockN{Artem~Polyvyanyy~\orcidartem}
\IEEEauthorblockA{
The University of Melbourne\\
Email: artem.polyvyanyy@unimelb.edu.au}
\and
\IEEEauthorblockN{Alistair~Moffat~\orcidalistair}
\IEEEauthorblockA{
The University of Melbourne\\
Email: ammoffat@unimelb.edu.au}
\and
\IEEEauthorblockN{Luciano~Garc{\'{\i}}a{-}Ba{\~{n}}uelos~\orciluciano}
\IEEEauthorblockA{
Tecnol\'ogico de Monterrey\\
Email: luciano.garcia@tec.mx}
}

\begin{document}

\maketitle


\begin{abstract}
Given an event log as a collection of recorded real-world process
traces, process mining aims to automatically construct a process
model that is both simple and provides a useful explanation of the
traces.
Conformance checking techniques are then employed to characterize and
quantify commonalities and discrepancies between the log's traces and
the candidate models.
Recent approaches to conformance checking acknowledge that the
elements being compared are inherently stochastic -- for example,
some traces occur frequently and others infrequently -- and seek to
incorporate this knowledge in their analyses.

Here we present an {\emph{entropic relevance}} measure for stochastic
conformance checking, computed as the average number of bits required
to compress each of the log's traces, based on the structure and
information about relative likelihoods provided by the model.
The measure penalizes traces from the event log not captured by the
model and traces described by the model but absent in the
event log, thus addressing both precision and recall quality criteria at the same time.
We further show that entropic relevance is computable in time linear
in the size of the log, and provide evaluation outcomes that
demonstrate the feasibility of using the new approach in industrial
settings.
\end{abstract}

\section{Introduction}
\label{sec:introduction}

\noindent
Process mining studies tools, methods, and techniques for improving
real-world processes based on event data generated by historical
process executions~\cite{Aalst2019}.
The core problem in process mining is that of automatically
discovering a process model from a given \emph{event log}, where an
event log is a collection of traces, each capturing a sequence of
observed process events.
Such discovered models should faithfully encode the process behavior
captured in the log and, hence, satisfy a range of criteria.
Specifically: (1) a discovered model should describe as many as
possible of the traces recorded in the log (good {\emph{recall}}, or
{\emph{fitness}}); (2) should allow as few traces as possible that
are not present in the log (good {\emph{precision}}); and (3) should
be as ``{\emph{simple}}'' as is consistent with the other two goals
(good {\emph{simplicity}}).
In tension with these three is a further objective: (4) the model
should allow traces that may stem from the same process but are not
present in the sample (good {\emph{generalization}}).
Conformance checking is a subarea of process mining that addresses
the problem of measuring and characterizing the four quality criteria
when using a discovered process model to explain the corresponding
event log.

Classical process mining techniques often consider frequencies of
traces in the logs and/or frequencies of events in the traces.
However, the implications of such considerations usually stay hidden
from the consumers of the artifacts that are produced.
For instance, most existing discovery algorithms strive to construct
process models that fit, {\ie} can replay, frequent traces from the
input event logs.
However, the vast majority of the discovery techniques construct
non-deterministic models in which routing decisions are equiprobable,
significantly limiting the ability of the models to explain the
behaviors of the true processes sampled via the event logs.
Indeed, if traces in a log suggest that failure was observed in nine
out of ten traces, a model that says that (only) every second trace
is erroneous is of reduced utility, and potentially harmful to
subsequent decision-making activities.
By accumulating event data over a long time, the log approaches the
true event and trace frequencies.
If this information about the true process were able to be
reflected in the discovered model, the model would generate better
process simulations, and hence better predictions of future processes.

We refer to process mining endeavors that process and produce
artifacts with explicit information on the likelihood of process
events and traces collectively as {\emph{stochastic}}, or
{\emph{statistical}}, {\emph{process mining}}.
For example, a discovery algorithm could construct a model with
annotations of relative likelihoods of making routing decisions such
that a large collection of traces induced by the model (by following
the encoded stochastic decisions) would result in a probability
distribution over traces that closely matches the distribution
of log traces.
We call such a model a {\emph{stochastic process model}}.
Before designing algorithms for discovering stochastic process models
from event logs, we seek to understand which stochastic models can be
considered to be ``good'' explanations of a given log.

Hence this paper, in which we present a novel technique for
stochastic conformance checking called {\emph{entropic
{\relevance}}}, or {\emph{\relevance}}.
Given a log, the entropic {\relevance} of a stochastic process
model is the average number of bits used to compress
{\cite{Cover2006}} a trace from the log using the relative
likelihoods induced by the model.
The fewer bits are used, the better the model ``explains'' the event
log, {\ie} the closer the relative likelihoods of traces derived from
the model to those obtained from the event log.
Log traces that do not fit the model are penalized by encoding them
using a more expensive background model.
Entropic {\relevance} also penalizes traces that fit the model but
are not present in the log, as these reduce the likelihoods of the
log traces that fit the model, and increase the length (in terms of
bits) of their compressed forms.
Hence, an entropic relevance measurement reflects a compromise
between the precision and recall quality criteria.
Such duality is indeed expected in the case of stochastic conformance
checking.
Finally, given a model and log, relevance is
computable in time linear in the size of the log, making it -- as we
demonstrate below -- useful when evaluating the quality of the
process discovery algorithms commonly used in the industry.

The remainder of the paper proceeds as follows.
The next section presents an overview of the compression methodology
we employ, and describes how an event log can be encoded via a
stochastic process model.
{\cref{sec:stochastic}} discusses formal stochastic models used in
process mining.
Based on these models, {\cref{sec:entropic}} presents the notion of
entropic relevance.
{\cref{sec:evaluation}} discusses the results of our evaluation of
the entropic relevance measure, while {\cref{sec:related:work}} summarizes related work.


\section{Motivation and Overview}
\label{sec:overview}

\noindent
We employ a minimum description length compression-based framework to
measure the quality of process models.
Two key observations make this possible: that a good process model is
one which accurately describes an observed set of traces, taken as a
sample from an underlying universe of traces; and that the
``describes'' operation can be precisely quantified by assessing the
cost of compressing the set of traces relative to the stochastic
language expressed by the model.
A signal benefit of this {\emph{entropic \relevance}} approach is that
not only is the model {\emph{structure}} an influence on its measured
usefulness, but also the {\emph{probability}} of each individual
trace having emerged from the model.
A relationship fundamental to information theory is critical to
understanding the new approach: if a symbol $e$ of probability $p(e)$
occurs, then the information conveyed by that occurrence is $-\log_2
p(e)$ bits, and hence that is also the minimum number of bits
required to describe any instance of~$e$.
For example, if $e$ has probability $p(e)=0.8$, then each $e$ that
occurs in a stream of such symbols has a cost of $0.3219$ bits
attributable to it.
Practical compression systems operate very close to these
entropy-based limits, see, for example, Moffat and Turpin
{\cite[Chapter 5]{mt02caca}}; and the information-theoretic
relationship between probabilities and bits is an achievable
one.

\fig{t}{0}{0.6}{0 0 0 0}{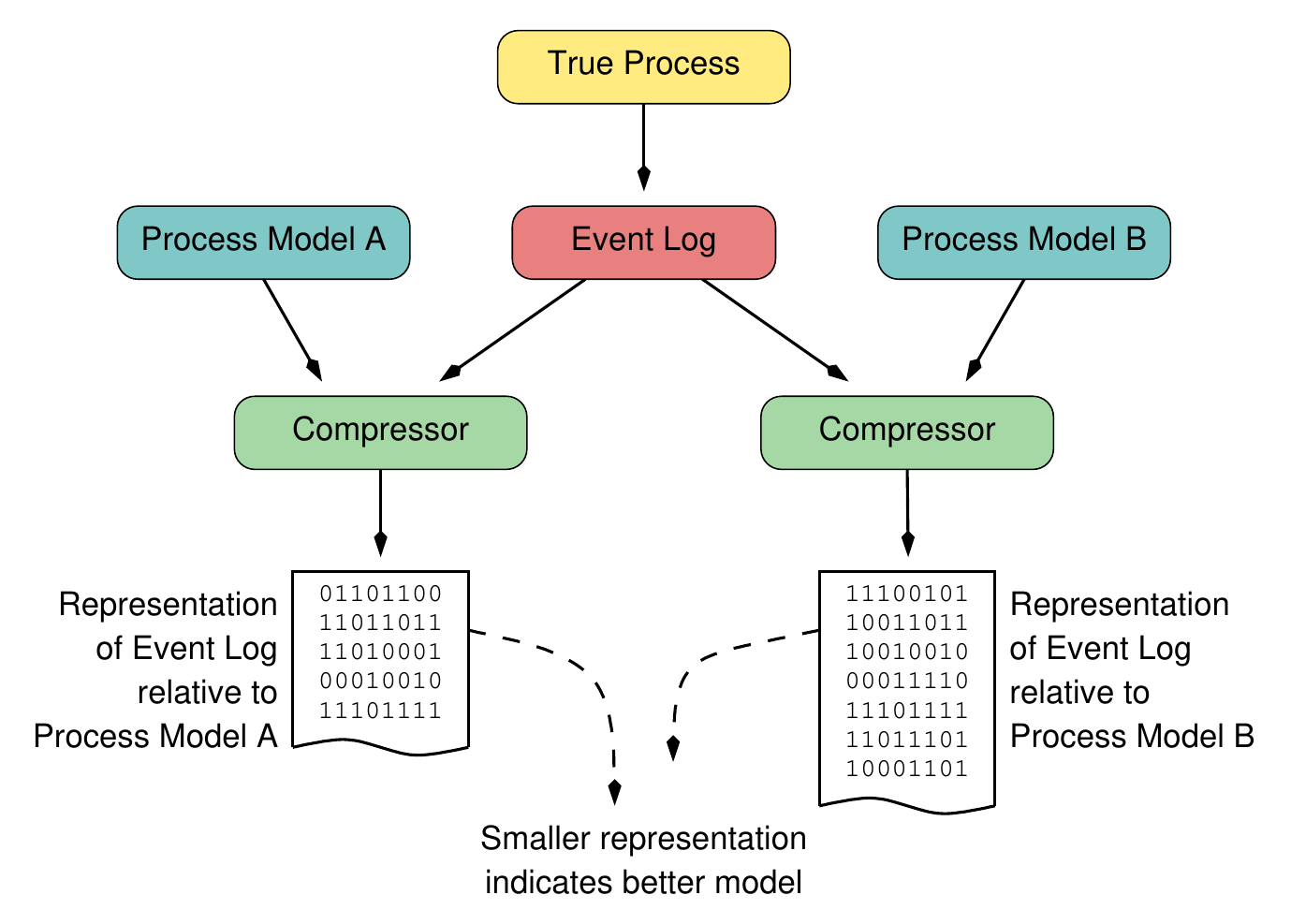}{-2}{
To measure the entropic relevance of a process model to a collection of traces, 
the model's structure and probabilities are used
to compute the cost in bits of losslessly representing the traces
relative to the model model.
Better models lead to a shorter compressed forms
}{-2}{fig:rel} 

{\cref{fig:rel}} provides an overview of our proposal.
In the figure it is supposed that an event log has been provided,
sampled from an underlying ``true'' (but unknown) process; and that
two process models are being considered as alternative explanations
for that set of observations.
If each of the two models assigns a calculable probability to every
possible sequence of process states that might occur, then computing
$\sum_{\strace} -\log_2 p(\trace\mid M)$ values, where the summation
is over the traces $t$ in the log and $p(\trace\mid M)$ is the
probability assigned to trace $\trace$ by model $M$, results in a
value that encapsulates the information content of the whole log,
given~$M$.
The smaller that value, the shorter the compressed output would be
were it to be generated, and the better the model matches the log.
(That is, while {\cref{fig:rel}} suggests that actual compression
occurs, it is the {\emph{size}} of the output that is of interest,
and not the actual bits that arise.)

\fig{t}{0}{0.6}{0 0 0 0}{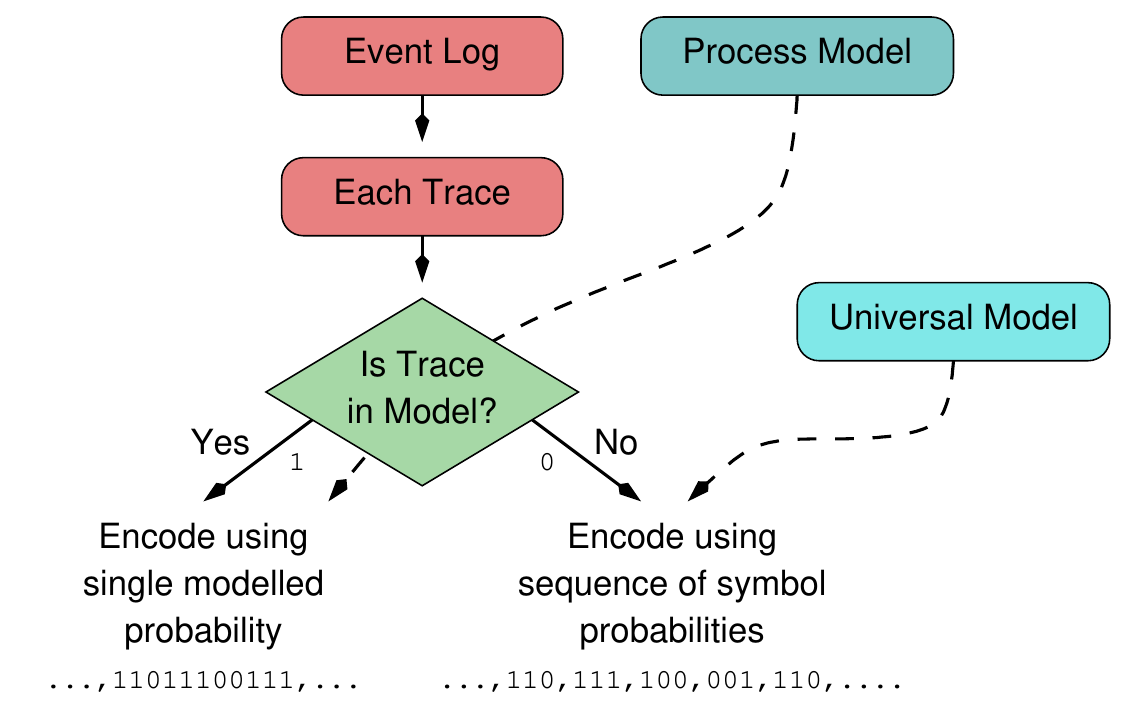}{-2}{
The two possible options when encoding a set of traces with respect
to a probabilistic model: each trace either has a non-zero
probability in the model, which can be used to derive a
bitstring; or it is spelled-out as symbol-by-symbol codes
via a universal model
}{-4}{fig:costs}

{\cref{fig:costs}} provides more details of the compression process,
considering each trace in the log.
The stochastic process model (defined in detail in
{\cref{sec:stochastic}}) assigns calculable non-zero probabilities to
a finite or infinite subset of the universe of possible traces,
rather than to every possible member of that universe; and hence
assigns a probability of zero to each of the infinite number of
possible state sequences in the complement subset.
If some particular trace in the log fits the model, it can be coded
as single entity, using its corresponding end-to-end probability in
the model.
On the other hand, the traces with a probability of zero according to
the model must be ``spelled out'' on a symbol-by-symbol basis, using
a background (or {\emph{universal}}) model in which every possible
state always has a non-zero probability, and hence in which every
possible sequence of states can always be coded.
To choose between these two cases, the output associated with every
trace is prefixed by a code -- a (biased) 0 or 1 bit -- that
indicates which option applies.
Given such an encoding and the stochastic and background models, one can always reconstruct the original log by applying the reverse procedure.

\fig{t}{-2}{0.7}{0 0 0 0}{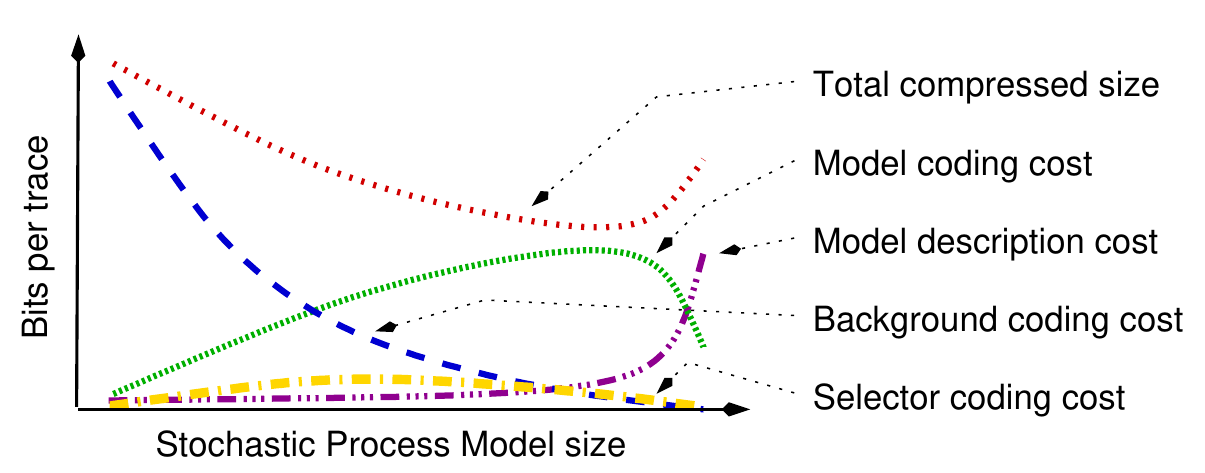}{-2}{
Schematic showing the total compressed size of a collection of traces
relative to a model as the sum of four components: the cost of
describing the model and its parameters; the cost of entropy coding
the traces that fit the model; the cost of entropy coding the traces
that do not fit the model, using a catch-all background technique;
and the cost of selecting, for each trace, which approach is used to
code it
}{-4}{fig:graphs} 

{\cref{fig:graphs}} steps back from the detail and provides a
high-level view of the proposed mechanism.
There are four components that collectively sum to the compressed
size, and that vary in different ways as the process model changes.
At the left end of the figure, if the process model is small and
easily described, it likely fits only a small fraction of the log's
traces.
The majority of traces, the ones that do not fit the model, are coded
using the more expensive background model, and it dominates to total
cost.
As the process model becomes larger and more sophisticated, it fits a
greater fraction of the traces, and the balance shifts from the
background model to the more economical stochastic process model.
The total compressed size decreases as this transition takes place.
Throughout this normal operating range the contributions of the other
two factors -- the binary per-trace selector flag, and the
description of the process model -- are typically very small
overheads.

In the limit, at the right of {\cref{fig:graphs}}, the process model
becomes large and is over-fitted to the traces in the particular
log.
The cost of using an over-fitted model is low, since each pathway
through it is unique; the cost of the background model is also low,
since no sequences need to be processed via it; and there is no cost
involved in selecting between the stochastic model and the background
model.
However the total compressed size will have increased, because of the
complexity and detail required in the description of the process
model.

In terms of {\cref{fig:graphs}}, we define (see {\cref{sec:entropic}}
for full details) entropic relevance to be the sum of the selector
coding cost, the background coding cost, and the model coding cost.
It is useful to retain the model size (measured in some conventional
manner) as a second dimension, shown as the horizontal axis in the
plot.
Furthermore, since process models are discrete objects (rather than a
continuous phenomena) the separation between entropic relevance and
model size allows definition of a Pareto frontier, {\ie} the set
of models that are either smaller in size, or superior in terms of
entropic relevance, to other possible models.

Entropic relevance (again, in anticipation of {\cref{sec:entropic}})
is measured in ``bits per trace'', with small values being preferable
to large ones.
The numeric range is open-ended, and it is neither desirable nor
possible to normalize the measurement in any way to obtain a ``0 to
1'' range.
Instead, it has meaningful units that clearly indicate the complexity
of the process that is being represented by the model.
With that understanding established, the process mining desiderata
listed at the beginning of {\cref{sec:introduction}} can be
considered: (1) traces not covered by the model must be coded using
the background predictions, increasing the net bit cost; (2)
processes permitted by the process model but not present in the log
cause the imputed probabilities of traces that do occur to decrease,
again increasing the net bit cost; and (3) simple models have smaller
model description costs, decreasing the net bit cost.
Objective (4) can also be accounted for, by noting that the
background model is always available, so hitherto unseen traces can
be accommodated, albeit with increased net bit costs.

\section{Models of Stochastic Languages}
\label{sec:stochastic}

\noindent
This section introduces the notion of a stochastic language and several models, theoretical and those used in practice, that aim at encoding stochastic languages.
The notion and models are used in the subsequent formal discussions and explanations of the conducted empirical evaluations.

\subsection{Stochastic Languages}
\label{subsec:stochastic:languages}

\noindent
A \emph{language} is a, possibly infinite, collection of finite
sequences of \emph{symbols}.
These sequences are often referred to as \emph{words}.
In this work, we use words to encode observed processes.
Hence, we refer to words as (process) \emph{traces} composed of
\emph{actions} rather than symbols.
Let $\actions$ be a universe of \emph{actions}.
Then, $\kleenestar{\actions}$ is the set of all traces over
$\actions$.
By $\emptysequence$, $\emptysequence \in \kleenestar{\actions}$, we
denote the empty trace.
For example, set
$X=\set{\emptysequence,\texttt{a},\texttt{ab},\texttt{abc},\texttt{abcd},\texttt{abcdd},\texttt{abcde}}$
defines a language of seven traces; we write $\texttt{abc}$ to denote
sequence $\sequence{\texttt{a},\texttt{b},\texttt{c}}$ when the
context is clear.

A {\emph{stochastic language}} is an assignment of probabilities to
traces so that the assigned probabilities sum up to one, {\ie} a
stochastic language is a probability density function over traces.
For example, a stochastic language might be used to encode the
relative likelihoods of observing words in a book, or encountering
traces in an event log of a software system.
A stochastic language is defined as follows.

\medskip
\begin{define}{Stochastic language}{def:stochastic:language}{\quad\\}
A \emph{stochastic language} $L$ is a function $\func{L}{\kleenestar{\actions}}{\intervalcc{0}{1}}$ for which it holds that:
\[
\sum_{\sigma \in \kleenestar{\actions}} \funcCall{L}{\sigma} = 1.0.
\]
\vspace{-2mm}
\end{define}
\medskip

\noindent
For example, $L_1=\set{\pair{\emptysequence}{0.5},\pair{\texttt{a}}{0.25},\pair{\texttt{ab}}{0.125},\pair{\texttt{abc}}{\nicefrac{1}{16}},\pair{\texttt{abcd}}{\nicefrac{1}{32}},\pair{\texttt{abcdd}}{\nicefrac{1}{64}},\pair{\texttt{abcde}}{\nicefrac{1}{64}}} \cup\, \smash{\bigcup_{\strace \in \kleenestar{\actions} \setminus X}{\set{\pair{\trace}{0.0}}}}$, where set $X$ is specified above.
Given a trace $\trace \in \kleenestar{\actions}$ and a stochastic
language $L$, $\funcCall{L}{\trace}$ specifies the relative likelihood of
a randomly drawn trace to be equal to $\trace$.

By $\hat{L}$, we denote the set of all traces possible according to
$L$, {\ie} $\hat{L} := \setbuilder{\trace \in
\kleenestar{\actions}}{\funcCall{L}{\trace}>0.0}$.
We say that $L$ is {\emph{finite}} if and only if $\hat{L}$ is finite;
otherwise $L$ is infinite.
It holds that $\hat{L}_1=X$, {\ie} the traces in $X$ are all the
possible traces according to $L_1$, and thus $L_1$ is finite.

\subsection{Stochastic Deterministic Finite Automata}

\noindent
A stochastic deterministic finite automaton (SDFA) can be used to
encode a stochastic language; here we adopt the definition of an SDFA
from Carrasco~\cite{Carrasco1997}.

\medskip
\begin{define}{Stochastic deterministic finite automaton}{def:SDFA}{\quad\\}
A \emph{stochastic deterministic finite automaton} (SDFA) is a tuple $(S, \Delta, \delta, p, s_0)$, where
$S$ is a finite set of \emph{states}, 
$\Delta \subseteq \actions$ is a set of \emph{actions}, 
$\func{\delta}{S \times \Delta}{S}$ is a \emph{transition function}, 
$\func{p}{S \times \Delta}{\intervalcc{0}{1}}$ is a \emph{transition probability function},
$s_0 \in S$ is the \emph{initial state}, and
for each state $s \in S$ it holds that 
${\sum_{\lambda \in \Delta} \funcCall{p}{s,\lambda}} \leq 1.0$.
\end{define}
\medskip

\noindent
By $\mathcal{A}$, we denote the universe of SDFAs.

\cref{fig:SDFA:0} shows an SDFA using graphical notation.
In this notation, the states and transition function are visualized as circles and arcs, respectively. 
For instance, the SDFA shown in \cref{fig:SDFA:0} has seven states $s_0 \ldots s_6$, and its transition function is defined by 
$\set{\tuple{s_0,\texttt{a},s_1},\tuple{s_1,\texttt{b},s_2},\tuple{s_2,\texttt{c},s_3},\tuple{s_3,\texttt{d},s_4},\tuple{s_4,\texttt{d},s_5},\tuple{s_4,\texttt{e},s_6}}$.
Arcs are labeled by actions and transition probabilities.
Hence, the arc from state $s_0$ to state $s_1$ with label ``$\texttt{a}(\nicefrac{1}{2})$'' specifies that $\tuple{s_0,\texttt{a},s_1} \in \delta$ and $\tuple{s_0,\texttt{a},0.5} \in p$.
Consequently, the transition probability function is defined by $\set{\tuple{s_0,\texttt{a},0.5},\tuple{s_1,\texttt{b},0.5},\tuple{s_2,\texttt{c},0.5},\tuple{s_3,\texttt{d},0.5},\tuple{s_4,\texttt{d},0.25},\tuple{s_4,\texttt{e},0.25}}$.
State $s_0$ is the initial state and, hence, is denoted by an arrow leading to it.

\begin{figure}[h!]
\begin{center}
\vspace{-1mm}
\begin{tikzpicture}[scale=0.8, transform shape, ->, >=stealth', shorten >=1pt, auto, initial text=, node distance=18mm, on grid, semithick, every state/.style={fill=orange!20, draw, circular drop shadow, text=black, minimum size=9mm}]
\node[initial,state,label=270:$s_0$]	(s0) 											{$\nicefrac{1}{2}$};
\node[state,label=270:$s_1$]		(s1) [right=of s0] 				{$\nicefrac{1}{2}$};
\node[state,label=270:$s_2$] 		(s2) [right=of s1]				{$\nicefrac{1}{2}$};
\node[state,label=270:$s_3$] 		(s3) [right=of s2]				{$\nicefrac{1}{2}$};
\node[state,label=270:$s_4$] 		(s4) [right=of s3]				{$\nicefrac{1}{2}$};
\node[state,label=0:$s_5$] 			(s5) [above right=of s4]	{$1$};
\node[state,label=0:$s_6$] 			(s6) [below right=of s4]	{$1$};

\path (s0) edge node {$\texttt{a}(\nicefrac{1}{2})$} (s1)
			(s1) edge node {$\texttt{b}(\nicefrac{1}{2})$} (s2)
			(s2) edge node {$\texttt{c}(\nicefrac{1}{2})$} (s3)
			(s3) edge node {$\texttt{d}(\nicefrac{1}{2})$} (s4)
			(s4) edge node {$\texttt{d}(\nicefrac{1}{4})$} (s5)
					 edge node {$\texttt{e}(\nicefrac{1}{4})$} (s6);
\end{tikzpicture}
\vspace{-2mm}
\caption{An SDFA.}
\label{fig:SDFA:0}
\vspace{-2mm}
\end{center}
\end{figure}

Given two traces $\trace_1,\trace_2\in\kleenestar{\actions}$, by $\concat{\trace_1}{\trace_2}$, we denote their concatenation, {\ie} the trace obtained by joining $\trace_1$ and $\trace_2$ end-to-end.
For example, it holds that $\concat{\texttt{tr}}{\texttt{ace}}=\texttt{trace}$.

An SDFA $A=(S, \Delta, \delta, p, s_0)$ encodes stochastic language $L_A$ defined using recursive function $\func{\pi_A}{S\times\kleenestar{\Lambda}}{\intervalcc{0}{1}}$, {\ie} $\funcCall{L_A}{\trace} := \funcCall{\pi_A}{s_0,t}$, $\trace \in \kleenestar{\actions}$, where:
\begin{equation*}\label{eq:sdfa:language}
\begin{split}
\funcCall{\pi_A}{s,\emptysequence} 				& := 1.0 - \sum_{\lambda \in \Delta} \funcCall{p}{s,\lambda}, \text{and}\\
\funcCall{\pi_A}{s,\concat{\lambda}{\trace'}}	& := \funcCall{p}{s,\lambda}\mult\funcCall{\pi_A}{\funcCall{\delta}{s,\lambda},t'}, \lambda \in \actions, t=\concat{\lambda}{\trace'}.
\end{split}
\end{equation*}

\noindent
Note that $\funcCall{\pi_A}{s,\emptysequence}$ denotes the probability of terminating a trace in state $s$ of $A$.
Such probabilities are shown diagrammatically as labels inside of the corresponding states.
For example, for SFDA $A$ from \cref{fig:SDFA:0}, it holds that $\funcCall{\pi_A}{s_i,\emptysequence}=0.5$, $i \in \intintervalcc{0}{4}$, and $\funcCall{\pi_A}{s_j,\emptysequence}=1.0$, $j=5$ or $j=6$.

If $L$ is a stochastic language encoded by some SDFA, we say that $L$ is a \emph{regular stochastic language}.
The SDFA in \cref{fig:SDFA:0} encodes stochastic language $L_1$ discussed in \cref{subsec:stochastic:languages} and, thus, $L_1$ is regular.

\subsection{Event Logs}
\label{subsec:event:logs}

\noindent
An \emph{event log} is a finite collection of events that relate to a process and are distinguished by their attributes and attribute values.
Usually, events in an event log encode information about actions executed by software systems that support a business process of an organization.
In general, an event can have arbitrary attributes, but three attributes are common in process mining. 
These are the \emph{case identifier}, \emph{timestamp}, and the \emph{action identifier} attribute.
The value of the case identifier attribute of an event relates this event to a case, or instance, of the process; {\ie} all events with the same case identifier stem from the same instance of the business process.
The time of occurrence of an event is stored in its timestamp attribute.
Finally, the action identifier attribute is used to store information about an action that induced the event.

In this work, we are neither interested in the exact times of event occurrences (but only in their orderings) nor in the distinctions between events and actions.
Thus, we encode all the events with the same case identifier as a trace of corresponding actions (obtained via the action identifier attribute) arranged in the ascending order of the event timestamps.
Finally, as there can be several case identifiers that induce the same trace (indeed, several business processes can induce the same sequence of actions with different timestamps), for our needs, it is convenient to represent an event log as a multiset of traces.

\medskip
\begin{define}{Event log}{def:event:log}{\quad\\}
An \emph{event log}, or \emph{log}, is a finite multiset of traces.
\end{define}
\medskip

\noindent
By $\mathcal{E}$, we denote the universe of logs.
For example, 
$E_1=\mset{\msetel{\emptysequence}{32}, \msetel{\texttt{a}}{16}, \msetel{\texttt{ab}}{8}, \msetel{\texttt{abc}}{4}, \msetel{\texttt{abcd}}{2}, \msetel{\texttt{abcdd}}{1}, \msetel{\texttt{abcde}}{1}}$
and
$E_2=\mset{\msetel{\emptysequence}{250},\msetel{\texttt{ab}}{250},\msetel{\texttt{abc}}{250},\msetel{\texttt{abcd}}{50},\msetel{\texttt{abce}}{50},\msetel{\texttt{abcde}}{50},\msetel{\texttt{abced}}{50},\msetel{\texttt{abcdde}}{50}}$
are two logs, {\ie} $E_1,E_2 \in \mathcal{E}$.
Trace $\texttt{abc}$ occurs in $E_1$ four times, while in $E_2$ it is recorded 250 times.

An event log is inherently stochastic.
By accumulating a large number of traces, and perhaps over an
extended period of time, an event log aims to approach their true
underlying probability distribution.
Let $X$ be a random variable and let $O$ be a multiset of
observations.
By $\funcCall{P}{X=x \mid O}$, or $\funcCall{P}{x \mid O}$ when the
context is clear, we denote the estimate based on $O$ of the
probability of observing $X$ to be equal to element $x$.
Given an event log $E \in \mathcal{E}$, we define the
\emph{stochastic language $L$ of $E$} by assigning probability
$\funcCall{L}{\trace}:=\funcCall{P}{\trace \mid E}$ to each trace
$\trace \in \kleenestar{\actions}$.
In this work, we use the maximum likelihood estimation, {\ie}
$\funcCall{P}{\trace \mid
E}:=\nicefrac{\multiplicity{E}{\trace}}{\cardinality{E}}$, where
$\multiplicity{E}{\trace}$ denotes the multiplicity of element
$\trace$ in multiset $E$.
Therefore, $L_1$ from \cref{subsec:stochastic:languages} is the stochastic language of event log $E_1$ from above.
Note that the stochastic language $L_2$ of event log $E_2$ is given my the function with these non-zero values
$\set{\pair{\emptysequence}{0.25},\pair{\texttt{ab}}{0.25},\pair{\texttt{abc}}{0.25},\pair{\texttt{abcd}}{0.05},\pair{\texttt{abce}}{0.05},\pair{\texttt{abcde}}{0.05},\pair{\texttt{abced}}{0.05},\pair{\texttt{abcdde}}{0.05}}.$

\subsection{Frequency Directed Action Graphs}
\label{sec:FDAG}

\noindent
A common approach for representing event logs for consumption and decision making by practitioners is by encoding them into Directly-Follows Graphs (DFGs)~\cite{Aalst2019}.
A DFG of an event log $E$ is a digraph in which vertices are actions encountered in the traces of $E$ and edges encode the directly-follows relation over the actions, {\ie} the DFG contains an edge directed from action $\texttt{a}$ to action $\texttt{b}$ \iffs $E$ contains a trace $\concat{\concat{\trace_1}{\texttt{ab}}}{\trace_2}$ where $\trace_1,\trace_2 \in \kleenestar{\actions}$~\cite{Aalst2016}.

As DFGs of industrial event logs are immense, they are often post-processed by filtering out vertices and edges that correspond, respectively, to infrequent actions and pairs of subsequent actions in the log. 
The vertices and edges of the filtered graphs are then annotated with numbers that reflect the frequencies of observing the corresponding concepts in the log. 
The frequencies aim to reflect the stochastic nature of the processes encoded in the corresponding log to inform the decision-making practices.
To describe such filtered graphs mathematically, we introduce the notion of a \emph{frequency directed action graph}.

\medskip
\begin{define}{Frequency directed action graph}{def:FDAG}{\quad\\}
A \emph{frequency directed action graph} (FDAG) is a tuple $(\Phi,\Psi,\phi,\psi,i,o)$, where 
$\Phi \subseteq \actions$ is a set of \emph{actions}, 
$\Psi \subseteq ((\Phi \times \Phi) \cup (\set{i} \times \Phi) \cup (\Phi \times \set{o}))$ is a \emph{directly-follows relation},
$\func{\phi}{\Phi \cup \set{i,o}}{\natnumwithzero}$ is an \emph{action frequency function},
$\func{\psi}{\Psi}{\natnumwithzero}$ is an \emph{arc frequency function},
and $i \not\in \actions$ and $o \not\in \actions$ are the input and the output of the graph, respectively.
\end{define}
\medskip

\noindent
By $\mathcal{G}$, we denote the universe of FDAGs.

\cref{fig:FDAG:0} shows an example FDAG. 
In the figure, boxes with rounded corners represent actions, whereas arcs encode the directly-follows relation.
The input node and the output node are denoted by $i$ and $o$, respectively.
The input node has no incoming arcs, while the output node has no outgoing arcs.
Finally, the action and arc frequencies assigned by the corresponding frequency functions are encrypted next to the respective actions and arcs. 

\begin{figure}[h!]
\begin{center}
\vspace{-1mm}
\begin{tikzpicture}[scale=0.7, transform shape, ->, >=stealth', shorten >=1pt, auto, node distance=18mm, on grid, semithick, action/.style={fill=purple!20, draw, rounded corners, minimum size=9mm, drop shadow}]
\node[action,label=180:\small 256]			(n0) 											{\Large $i$};
\node[action,label=90:\small 192]			(n1) [right=of n0] 				{\Large $\texttt{a}$};
\node[action,label=90:\small 96] 			(n2) [right=of n1]				{\Large $\texttt{b}$};
\node[action,label=105:\small 48] 		(n3) [right=of n2]				{\Large $\texttt{c}$};
\node[action,label=270:\small 12] 		(n4) [right=of n3]				{\Large $\texttt{d}$};
\node[action,label=0:\small 18] 		(n5) [right=of n4]				{\Large $\texttt{e}$};
\node[action,label=270:\small 256] 		(n6) [below=of n3]				{\Large $o$};

\path (n0) edge node {\small 192} (n1)
					 edge [bend right=15] node {\small 64} (n6)
			(n1) edge node {\small 96} (n2)
					 edge [bend right=10] node {\small 96} (n6)
			(n2) edge node {\small 48} (n3)
					 edge node {\small 48} (n6)
			(n3) edge node {\small 12} (n4)
					 edge [bend left=85] node {\small 12} (n5)
					 edge node {\small 24} (n6)
			(n4) edge [loop above] node {\small 3} (n4)
					 edge node {\small 6} (n5)
					 edge node {\small 6} (n6)
			(n5) edge [bend left=15] node {\small 18} (n6)
;
\end{tikzpicture}
\caption{An FDAG.}
\label{fig:FDAG:0}
\vspace{-2mm}
\end{center}
\end{figure}

Van der Aalst {\cite{Aalst2019}} observes that practitioners can
interpret FDAGs in different ways.
Because of the filtering step, it is possible to associate a given
FDAG with different collections of traces.
He then discusses several pitfalls this phenomenon can lead to in
practice.
We agree with those observations, and, for our purpose, fix one such
possible interpretation.
To avoid ambiguities, next, we define our interpretation rigorously
as a mapping from a given FDAG to the corresponding SDFA.

\medskip
\begin{define}{SDFA of FDAG}{def:FDAG2SDFA}{\quad\\}
Let $G:=(\Phi,\Psi,\phi,\psi,i,o)$ be an FDAG.
Then, $(S,\Delta,\delta,p,s_0)$, where
$S = \Phi \cup \set{i}$, 
$\Delta = \Phi$, 
$\delta = \setbuilder{(s,t,t) \in (\set{i} \cup \Phi)\times\Phi\times\Phi}{(s,t) \in \Psi}$, 
$p = \setbuilder{(s,t,x) \in (\set{i} \cup \Phi)\times\Phi\times\intervalcc{0}{1}}{(s,t) \in \Psi \land x = \funcCall{\psi}{s,t} / \sum_{(s,u) \in \Psi}{\funcCall{\psi}{s,u}}}$, and
$s_o = i$,
is the SDFA of $G$, denoted by $\funcCall{SDFA}{G}$.
\end{define}
\medskip

\noindent
Thus, if $A=\funcCall{SDFA}{G}$, then, according to our interpretation, $G$ encodes the possible traces of $L_A$, {\ie} traces $\hat{L}_A$, and the relative frequency of each trace $\trace \in \hat{L}_A$ is given by $\funcCall{L_A}{\trace}$.

\begin{figure}[t]
\begin{center}
\begin{tikzpicture}[scale=0.8, transform shape, ->, >=stealth', shorten >=1pt, auto, initial text=, node distance=18mm, on grid, semithick, every state/.style={fill=orange!20, draw, drop shadow, text=black, minimum size=9mm}]
\node[initial,state,label=270:$q_0$]	(q0) 								{$\nicefrac{1}{4}$};
\node[state,label=270:$q_1$]		(q1) [right=of q0] 				{$\nicefrac{1}{2}$};
\node[state,label=270:$q_2$] 		(q2) [right=of q1]				{$\nicefrac{1}{2}$};
\node[state,label=260:$q_3$] 		(q3) [right=of q2]				{$\nicefrac{1}{2}$};
\node[state,label=270:$q_4$] 		(q4) [right=of q3]				{$\nicefrac{2}{5}$};
\node[state,label=0:$q_5$] 			(q5) [right=of q4]				{$1$};

\path (q0) edge node {$\texttt{a}(\nicefrac{3}{4})$} (q1)
			(q1) edge node {$\texttt{b}(\nicefrac{1}{2})$} (q2)
			(q2) edge node {$\texttt{c}(\nicefrac{1}{2})$} (q3)
			(q3) edge node {$\texttt{d}(\nicefrac{1}{4})$} (q4)
					 edge [bend right=445] node {$\texttt{e}(\nicefrac{1}{4})$} (q5)
			(q4) edge [loop above] node {$\texttt{d}(\nicefrac{1}{5})$} (q4)
					 edge node {$\texttt{e}(\nicefrac{2}{5})$} (q5)
			;
\end{tikzpicture}
\vspace{-4mm}
\caption{An SDFA.}
\label{fig:SDFA:1}
\vspace{-5mm}
\end{center}
\end{figure}
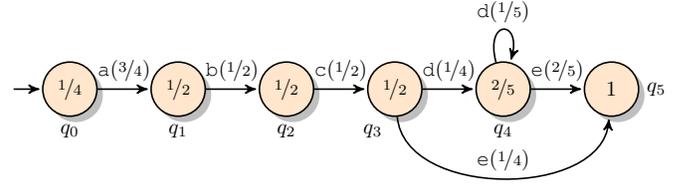

For example, \cref{fig:SDFA:1} shows the SDFA of the FDAG from \cref{fig:FDAG:0} that encodes infinite stochastic language $L_2$ which, among others, assigns these non-zero values to traces 
$\set{\pair{\emptysequence}{\nicefrac{1}{4}},\pair{\texttt{a}}{\nicefrac{3}{8}},\pair{\texttt{ab}}{\nicefrac{3}{16}},\pair{\texttt{abc}}{\nicefrac{3}{32}},\pair{\texttt{abcd}}{\nicefrac{3}{160}},\pair{\texttt{abce}}{\nicefrac{3}{64}},\pair{\texttt{abcdd}}{\nicefrac{3}{800}},\pair{\texttt{abcde}}{\nicefrac{3}{160}},\pair{\texttt{abcddd}}{\nicefrac{3}{4000}}, \pair{\texttt{abcdde}}{\nicefrac{3}{800}}}$; all the other possible according to $L_2$ traces have the cumulative relative frequency of $\nicefrac{9}{8000}$.

\section{Entropic {\Relevance}}
\label{sec:entropic}

\noindent
This section gives a precise definition of {\emph{entropic {\relevance}}}.
Given an event log $E \in \mathcal{E}$ and an SDFA $A$, trace
$\trace\in E$ is either a possible trace according to the stochastic
language of $A$, or is not.
This distinction determines how the number of bits required to encode
$\trace$ is computed.
If $\trace$ is possible, then it is presumed to be encoded using the
probability of $\trace$ according to $A$, {\ie}
$\funcCall{L_A}{\trace}$.
Otherwise, $\trace$ is presumed to be encoded using background
knowledge about the log.
These two modes are captured in this next definition.

\medskip
\begin{define}{Trace compression
cost}{def:trace:compression:cost}{\quad\\} Let $\trace\in E$ be a
trace in an event log $E\in\mathcal{E}$, let $A\in\mathcal{A}$ be an
SDFA, and let $\func{bits}{\kleenestar{\actions} \times \mathcal{E}
{\times} {\mathcal{A}}}{\nonnegrealnumbers}$ be a function that maps
any and every sequence $\trace\in\kleenestar{\actions}$ to the number
of bits required to uniquely encode $\trace$, given a viable encoding
that assumes knowledge of $E$ and $A$.
The {\emph{trace compression cost of $\trace$ in the presence
of $E$ and $A$}} is defined:
\[
\funcCall{cost_{bits}}{\trace,E,A}:=
\begin{cases} 
-\funcCall{\log_2}{L_A(\trace)} & t \in \hat{L_A} \\
\funcCall{bits}{\trace,E,A} & \text{otherwise.}
\end{cases}
\]
\end{define}
\medskip

For simplicity, in this first presentation we use a straightforward
background model to measure $\funcCall{bits}{\trace,E,A}$, ignoring
both $E$ and $A$, and trivially encoding $\trace$ by taking each
individual action as an equi-probable symbol over the underlying
alphabet augmented by an ``end of string'' symbol, and with each
trace terminated by that special symbol:
$\funcCall{bits}{\trace,E,A}:=(1+\seqLength{\trace})
\mult
\funcCall{\log_2}{1+\setCardinality{\actions}}$.
(We anticipate future exploration of more sophisticated encoding
schemes that explore partial embeddings of traces into finite-context
automata, and/or make use of the relative frequencies of the actions
present in~$E$.)

Given that definition, let $A_1$ and $A_2$ be SDFAs shown in
{\cref{fig:SDFA:0}} and {\cref{fig:SDFA:1}}, respectively.
Then, based on logs $E_1$ and $E_2$ from
\cref{subsec:event:logs}, the compression costs of traces
$\texttt{abcd}$ and $\texttt{abce}$ are (in\,{\bits}):

\smallskip
{\footnotesize
\begin{compactenum}
\vspace{.05mm}
\item \!\!\!\!$\funcCall{cost_{bits}}{\texttt{abcd},E_1,A_1}\!=\!-\funcCall{\log_2}{\funcCall{L_{A_1}}{\texttt{abcd}}}\!=\!-\funcCall{\log_2}{\nicefrac{1}{32}}\!=\!5.00$;
\vspace{.05mm}
\item \!\!\!\!\!$\funcCall{cost_{bits}}{\texttt{abcd},E_1,A_2}\!=\!\!-\funcCall{\log_2}{\funcCall{L_{A_2}}{\texttt{abcd}}}\!=\!-\funcCall{\log_2}{\nicefrac{3}{160}}\!\!\!=\!5.74$;
\vspace{.05mm}
\item \!\!\!\!\!\!$\funcCall{cost_{bits}}{\texttt{abce},E_2,A_1}\!\!=\!\!(1\!\!+\!\!\seqLength{\texttt{abce}}) \mult \funcCall{\log_2}{1\!+\!\setCardinality{\set{\texttt{a},\texttt{b},\texttt{c},\texttt{d},\texttt{e}}}}\!=\!12.93$;
\vspace{.05mm}
\item \!\!\!\!$\funcCall{cost_{bits}}{\texttt{abce},E_2,A_2}\!=\!-\funcCall{\log_2}{\funcCall{L_{A_2}}{\texttt{abce}}}\!=\!-\funcCall{\log_2}{\nicefrac{3}{64}}\!=\!4.42$.
\end{compactenum}
}
\smallskip
\noindent
Hence, $A_1$ compresses trace {\texttt{abcd}} better than does $A_2$,
while $A_2$ compresses {\texttt{abce}} much better than $A_1$.
In particular, {\texttt{abce}} is impossible according to
$A_1$, and is encoded using the $\funcCall{bits}{\cdot,\cdot,\cdot}$
function (case~3).
That encoding needs five symbols: four actions and an ``end
of string'' symbol, each taking $\log_2{6}$ bits because the alphabet in
$E_2$ contains five symbols, and an ``end of string'' symbol must also
be included.

Also needed in the nominal compression cost is the ``selector''
associated with the diamond decision box in {\cref{fig:costs}}, and
illustrated by the yellow line in {\cref{fig:graphs}}.
If the probability associated with a two-choice event is $p$, the
expected cost of coding a stream of such choices is given by
$\funcCall{H_0}{p} := -p \mult \funcCall{\log_2}{p} - \left( 1 - p
\right) \mult
\funcCall{\log_2}{1-p}$; with, by definition,
$\funcCall{H_0}{0.0}:=\funcCall{H_0}{1.0}:=0.0$.
These considerations lead directly to our main definition.

\medskip
\begin{define}{Entropic \relevance}{def:relevance}{\quad\\}
Let $E \in \mathcal{E}$ be an event log and let $A \in \mathcal{A}$
be an SDFA.
Let $\funcCall{\rho}{E,A}$ be the overall probability that a trace
in $E$ is possible in the stochastic language of $A$,
$\funcCall{\rho}{E,A} = \sum_{\trace \in
\hat{L_A}}{\funcCall{P}{\trace\mid E}}$.
Then, the {\emph{entropic {\relevance} of $A$ to $E$}}, or
{\emph{{\relevance} of $A$ to $E$}}, is denoted by
$\funcCall{rel}{E,A}$ and defined as:
\[
\funcCall{rel}{E,A} := H_0(\funcCall{\rho}{E,A})
	+ \frac{1}{\cardinality{E}} \sum_{\strace \in E}
		\funcCall{cost_{bits}}{\trace,E,A} \, .
\]
\end{define}
\medskip

{\cref{fig:costs}}, presented earlier, explains
{\cref{def:trace:compression:cost}} and {\cref{def:relevance}}.
During the nominal encoding process, each trace $\trace$ in the log
is considered in turn.
If $\trace$ has a non-zero probability in the model, the
corresponding selector code is generated (the left branch out of the
decision diamond in {\cref{fig:costs}}), and then that probability is
used to encode $\trace$ relative to the model (the first option in
{\cref{def:trace:compression:cost}}).
If the probability of $\trace$ is zero according to the model (the
right branch in {\cref{fig:costs}}), the opposite selector code is
needed, and then $\funcCall{bits}{\trace,E,A}$ bits are used to
represent $\trace$ on a symbol-by-symbol basis in the universal
background model (the second option in
{\cref{def:trace:compression:cost}}).
The selector codes associated with the diamond decision box, needed
to differentiate between the two alternatives on a per-trace basis,
add an average of $H_0(\funcCall{\rho}{E,A})$ bits per trace.

Again, consider automata $A_1$ and $A_2$ from {\cref{fig:SDFA:0}} and
{\cref{fig:SDFA:1}}, respectively, and event logs $E_1$ and $E_2$
from {\cref{subsec:event:logs}}.
{\cref{tab:running:examples}} summarizes the constituent costs and
the resulting entropic {\relevance} values for the four combinations.
In the first row, SDFA $A_1$ explains event log $E_1$ perfectly, as
the traces in the log all fit the automaton, and the relative
likelihoods of observing the traces in the log and in the automaton
are identical.
No other automaton can achieve lower {\relevance} for $E_1$, and
$A_1$ is {\emph{optimal}}.

Automaton $A_2$ explains $E_1$ reasonably well, but with an increased
model coding cost because of mis-matched probabilities
for the empty trace and for the one-action trace $\texttt{a}$
($\nicefrac{1}{2}$ vs $\nicefrac{1}{4}$ for the empty trace, and
$\nicefrac{1}{4}$ vs $\nicefrac{3}{8}$ for $\texttt{a}$).

\begin{table}[t]
\centering
\begin{tabular}{|c|c|c|c|c|c|c|}
\hline
Autom.	& Log	& $\rho$ & Select. & Bckgrd. & MdlCst. & \Relevance\ \\
\hline
\hline
$A_1$ &	$E_1$ &	1.00 & 0.00 &	0.00 & 1.97 & 1.97 \\
$A_2$ &	$E_1$ &	1.00 & 0.00 &	0.00 & 2.26 & 2.26 \\
$A_1$ &	$E_2$ &	0.85 & 0.61 &	2.33 & 2.55 & 5.49 \\
$A_2$ &	$E_2$ &	0.95 & 0.29 &	0.78 & 3.15 & 4.22 \\
\hline
\end{tabular}
\caption{
\small
Entropic {\relevance} (in {\bits}) and its constituents for
two example automata
and two event logs.
The columns list the probability of traces from the log being
possible according to the automata ($\rho$); and the average (over
all that log's traces) selector coding cost (Select.), background
coding cost (Bckgrd.), and model coding cost (MdlCst.).
The final column sums those three to get the entropic {\relevance}.}
\label{tab:running:examples}
\vspace{-3mm}
\end{table}

The probability $\funcCall{\rho}{E_2,A_1}$ that a trace from $E_2$ is
possible according to $L_{A_1}$ is $0.85$, and hence
$H_0(\funcCall{\rho}{E,A})=0.61\, {\bits}$.
That is, the choice between the background model and $A_1$ adds
$-\funcCall{\log_2}{\rho} = 0.23\,{\bits}$ to the traces in $E_2$ that
are possible according to $L_{A_1}$; adds
$-\funcCall{\log_2}{1-\rho} = 2.74\,{\bits}$ for traces that are not
possible; and averages at $0.61$ bits per trace.
The {\relevance} of $A_1$ to $E_2$ is then obtained by adding in the
arithmetic mean of the trace compression costs of the traces in
$E_2$.
The smallest compression cost arises for the empty trace
($2.59\,{\bits}$, part of the model coding cost), while the highest
cost is associated with $\texttt{abcdde}$ ($18.09\,{\bits}$, part of
the background coding cost).
Overall, $A_1$ can be used to compress $E_2$ using, on average,
$5.49\,{\bits}$ per trace.

In the last row of the table, $A_2$ ``explains'' $E_2$ using
$4.22\,{\bits}$ per trace, and hence $A_2$ is a better model for
$E_2$ than is~$A_1$.
First, note that $A_2$ fits more of $E_2$'s traces than does $A_1$,
and with $\funcCall{\rho}{E_2,A_1}=0.85$ and
$\funcCall{\rho}{E_2,A_2}=0.95$, the average cost of
selecting between the process model and background model is less for
$A_2$ than for $A_1$.
In direct correspondence, the share of {\relevance} stemming from the
background coding cost is also less for $A_2$ than for $A_1$.
Finally, despite the fact that more traces from $E_2$ fit $A_2$ than
$A_1$, which pushes $A_2$'s model coding cost higher, the overall
cost of using $A_2$ is less.

\section{Evaluation}
\label{sec:evaluation}

\newcommand{\company}	{{Celonis SE}}

\begin{figure*}[t]
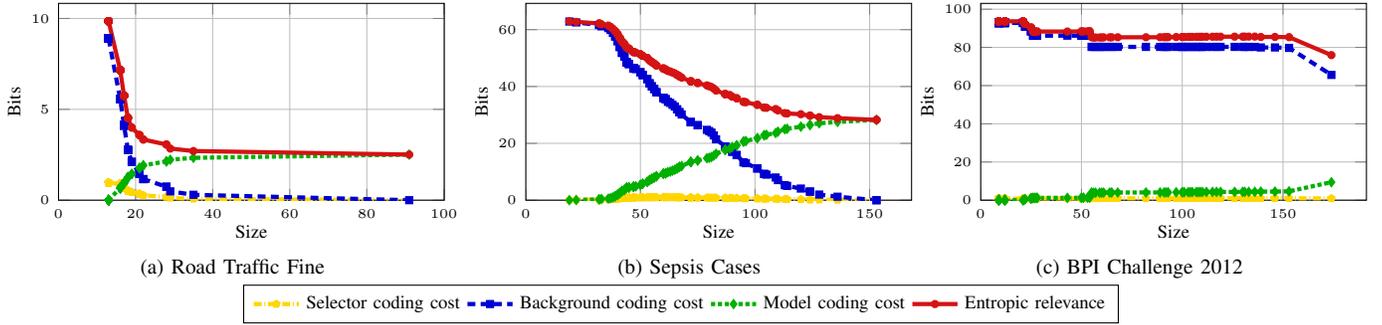

\vspace{-1.5mm}
\centering
\subfloat[Road Traffic Fine]{\relplot{data/sander/rel-traffic-fines.csv}%
\label{fig:plot:rel:fines}}%
\subfloat[Sepsis Cases]{\relplot{data/sander/rel-sepsis.csv}%
\label{fig:plot:rel:sepsis}}%
\subfloat[BPI Challenge 2012]{\relplot{data/sander/rel-financial.csv}%
\label{fig:plot:rel:financial}}%
\vspace{.5mm}
\ref{rellegend}
\caption{Entropic relevance and its constituents, plotted as a
function of model size measured as states plus edges.}
\label{fig:plots:rel}
\end{figure*}

\begin{figure*}[t]
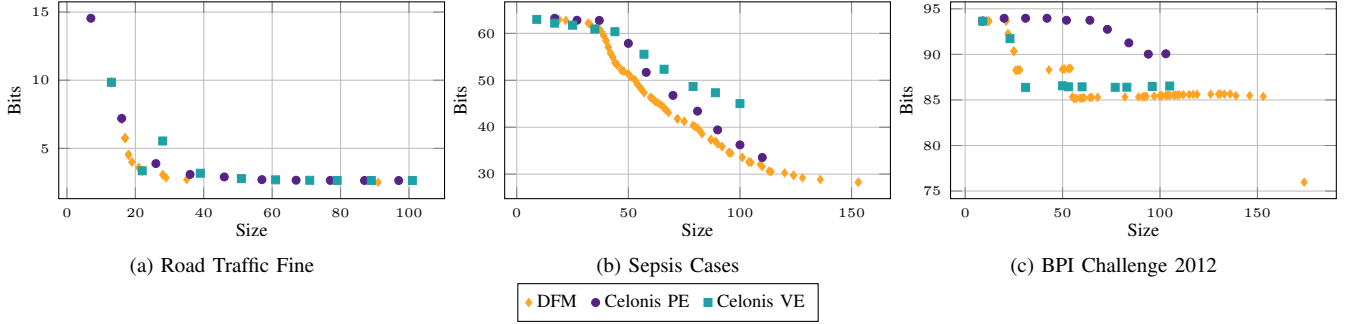

\vspace{-6mm}
\centering
\subfloat[Road Traffic Fine]{\scatterplot{data/celonis/rel-traffic-fines-pe.csv}{data/celonis/rel-traffic-fines-ve.csv}{data/sander/rel-traffic-fines.csv}%
\label{fig:plot:scatter:traffic:fines}}%
\subfloat[Sepsis Cases]{\scatterplot{data/celonis/rel-sepsis-pe.csv}{data/celonis/rel-sepsis-ve.csv}{data/sander/rel-sepsis.csv}%
\label{fig:plot:scatter:sepsis}}%
\subfloat[BPI Challenge 2012]{\scatterplot{data/celonis/rel-financial-pe.csv}{data/celonis/rel-financial-ve.csv}{data/sander/rel-financial.csv}%
\label{fig:plot:scatter:financial}}%
\vspace{.5mm}
\ref{relscatterlegend}
\caption{Entropic relevance, again plotted as a function of model
size, for three different techniques.}
\label{fig:plots:scatter}
\vspace{-4.5mm}
\end{figure*}

\noindent
To study the usefulness of entropic {\relevance}, we used
event logs from real-world IT-systems made publicly available by the IEEE
Task Force on Process
Mining.\footnote{\url{https://data.4tu.nl/repository/collection:event\_logs\_real}.}
For each log, directly follows models (DFMs, also known as
DFGs) were constructed using the approach of Leemans {\etal}
{\cite{Leemans2019b}}, with trace removal thresholds of
{\nicefrac{1}{100}}, {\nicefrac{2}{100}}, {\ldots},
{\nicefrac{100}{100}}.
{\cref{fig:plots:rel}} plots {\relevance} values (and
constituents)
for three logs: Road Traffic Fine
Management~\cite{DeLeoniM.Massimiliano2015}, Sepsis
Cases~\cite{Mannhardt2016a}, and BPI Challenge
2012~\cite{VanDongen2012}.
The {\emph{size}} of each model (the horizontal axis) is taken to be
the number of states plus the number of edges.

For the first two logs, the curves correspond to the anticipation
provided by {\cref{fig:graphs}}.
In both cases as the models become more complex, an increasing
fraction of traces fit the model (green line); a decreasing fraction
are coded using the background model (blue line); and the selector
coding cost (yellow line) is small over most of the range.
Entropic {\relevance} settles at around $2.5$ bits per trace for
Traffic Fines, indicating that the log is highly regular and hence
highly compressible; and at around $30$ bits per trace for Sepsis.

In the third log in {\cref{fig:plots:rel}}, a different picture
emerges.
Now the background model dominates the entropic cost and relevance
values are high, a consequence of long traces that tend not to repeat
in the log.
Changing to a more nuanced background model -- based on action
occurrence frequencies and a zero-order model, for example -- would
lower the relevance values, but not decrease the model's reliance on
them.
In the context of the framework introduced here, the traces in this
log are relatively inconsistent, and hence incompressible.

{\company} (\url{https://www.celonis.com}) granted us access to DFGs they constructed using ``Celonis Snap'', the free version of their enterprise-grade product.
For each of the logs in~\cref{fig:plots:rel}, they generated twenty DFGs using two different techniques (PE and VE) and
ten configurations for each technique. 
These DFGs were transferred to us on July 6, 2020.
{\cref{fig:plots:scatter}} plots relevance values in bits for these
DFGs, again as a function of model size, plus the DFMs already used
in {\cref{fig:plots:rel}}.
Now models, and hence methods, can be compared by considering the
Pareto frontier of the points plotted in each graph.
For these logs the relevance approach
identifies the methods of {\company} as being preferable when the
goal is to have small- to moderate-sized DFGs.
However, note that the numeric relevance values are subject to the
approach used to map DFGs to SDFAs and to the choice of background
coding model.
Detailed analysis of these interactions and of the DFGs discovered
using other techniques is future work.

The computation of entropic {\relevance} requires straightforward
data structures and execution loops.
With a hash-map used to implement the set of edges at each state in
the model, computation time is linear in the total volume of log data
processed (number of traces times average length). 
The average CPU time for computing relevance using our implementation on a commodity laptop computer for the largest analyzed log (BPI Challenge 2018 -- Payment application; 43,809 traces and 984,613 events) over the 100 constructed DFMs (size ranged from 25 to 238) was 0.47 sec.
Note also that none of the models plotted in \cref{fig:plots:rel,fig:plots:scatter} were
over-fitted to the data, and the model description costs (see
{\cref{fig:graphs}}) were small compared to the entropic relevance
scores.
Our tool~\cite{Polyvyanyy2020tool} and dataset~\cite{Alkhammash2020} used to conduct the experiments are publicly available.

\section{Related Work}
\label{sec:related:work}

\noindent
A plethora of non-stochastic process discovery and conformance
checking techniques have been proposed over the last two decades,
including the Genetic Mining~\cite{Medeiros2007}, Heuristic
Mining~\cite{Weijters2003}, Inductive Mining~\cite{Leemans2013}, and
Split Mining~\cite{AugustoCDRP18} algorithms, all of which have been
well-received by the process mining community.
Non-stochastic conformance checking techniques can be broadly
classified into {\emph{quantitative}}, those that summarize
conformance diagnostics into a single number, and
{\emph{qualitative}}, those that construct detailed analytics of
commonalities and discrepancies between model and log traces.
Carmona {\etal} {\cite{Carmona2018}} provide a useful overview.

Recently, Van der Aalst {\etal} {\cite{Aalst2018,Syring2019}}
initiated discussion of desired properties for conformance checking
techniques.
Entropy-based measures are the only quantitative conformance checking
techniques that are known to satisfy all the properties for precision
and recall that have been proposed to date {\cite{Polyvyanyy2020}}, 
including the strict monotonicity properties.

The selection of currently available stochastic process mining
techniques is rather scarce.
To the best of our knowledge, there are two stochastic discovery
techniques proposed by academia.
The technique presented by Rogge-Solti {\etal}
{\cite{RoggeSolti2013}} discovers stochastic Petri nets, while that
of Leemans {\etal} {\cite{Leemans2019b}} can be used to discover
DFGs, and was employed in {\cref{sec:evaluation}}.
There are many commercial tools for discovering DFGs, or FDAGs, from
event logs, but these are all closed source.

Two stochastic conformance checking techniques have been proposed.
Leemans {\etal} {\cite{LeemansSA19}} base their approach on the
``earth movers' distance'', and measure the effort to transform the
distribution of log traces into the distribution of model traces,
seen as two piles of dirt that need to be aligned with minimal
effort.
The technique is computationally demanding and suggests
practical trade-offs between accuracy, run time, and memory usage.
The approach of Leemans and Polyvyanyy {\cite{LeemansP20}} is
inspired by entropy-based conformance checking
{\cite{Polyvyanyy2020}}.
Leemans and Polyvyanyy {\cite{LeemansP20}} also suggest a range of
desired properties for stochastic precision and recall and
show that their measures indeed possess these properties.
The calculation of the measures requires (in the worst case) a
quadratic number of steps in the size of the corresponding SDFAs,
while entropic relevance runs in linear time in the size of the log.
In addition, relevance, as defined here, reflects the compromise
between precision and recall in a single value with meaningful units.
Exploration of the relationships and correlations between these two
previous measures and ours, and understanding of their differences,
is an area for future work.

Finally, note that our proposal is an application of the
{\emph{minimum description length}} principle
{\cite{bry98ieeeinfo,hy01jasa}}; which, in turn, is related to the
1965 definition by Kolmogorov that the intrinsic descriptive
complexity of an object is the length of the shortest binary computer
program that describes it~\cite{Cover2006}.
Kolmogorov complexity formalizes the notion widely known as ``Occam's
Razor'' {\cite{Tornay1938}}, a problem-solving principle attributed
to William of Ockham which suggests that the simplest (that is,
shortest) sufficient explanation of a phenomenon is the best.

\section{Conclusion}
\label{sec:conclusion}

\noindent
We have presented an entropic {\relevance} measure for stochastic
conformance checking.
The new measure is grounded in a minimum description length
compression-based framework that assesses how accurately a stochastic
process model describes an event log by computing the length of an
encoding of the log traces relative to the stochastic language
expressed by the model.
The relevance of a model to a given log reflects a compromise between
the precision and recall quality criteria, and is computable in time
linear in the size of the log.
Relevance is measured in bits, with values being directly
interpretable, and with small scores being preferable.

Future work will investigate the effects of using different
background models for calculating entropic relevance, with the aim of
identifying models that lead to useful relevance measurements; noting
that background model cost is one of the three components of the
relevance calculation, and in some cases is dominant.
We also plan to explore new techniques for discovering stochastic
process models in a direct response to entropic relevance.
Now that we have defined entropic {\relevance} as a useful quantity,
an important next step is to explicitly design models that seek to
minimize it.

\smallskip
\noindent
\textbf{Acknowledgment.}
Artem Polyvyanyy was in part supported by the Australian Research
Council project DP180102839.
Hanan Alkhammash and Thomas Vogelgesang provided assistance with the preparation of
the datasets used in {\cref{sec:evaluation}}.


\bibliography{bibliography}
\end{document}